\documentclass[11pt]{article}

\PassOptionsToPackage{table}{xcolor} 
\usepackage[final]{acl}

\usepackage{times}
\usepackage{latexsym}

\usepackage[T1]{fontenc}

\usepackage[utf8]{inputenc}

\usepackage{microtype}

\usepackage{inconsolata}

\usepackage{graphicx}

\usepackage{xspace}
\usepackage{fancyvrb}
\usepackage{enumitem}
\usepackage{wrapfig}
\usepackage[most]{tcolorbox} 
\usepackage{booktabs}      
\usepackage{array}         
\usepackage{caption}       
\usepackage{multirow}      
\usepackage{makecell}
\usepackage{amsmath} 
\usepackage{pifont}
\usepackage[misc]{ifsym}
\usepackage{amssymb}
\usepackage{tabularx}
\usepackage[linesnumbered,ruled,lined]{algorithm2e}
\usepackage{hyperref}
\usepackage{fontawesome5}   
\usepackage{graphicx}       
\newcommand{\hficon}{%
  \raisebox{-0.2\height}{\includegraphics[height=1em]{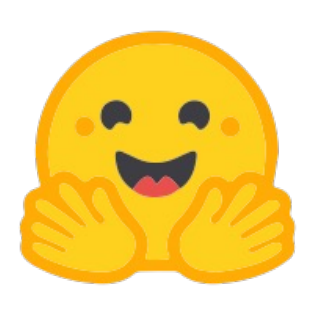}}%
}

\title{\ourbench: Can Language Models Answer Repository-level Code Questions?}

\author{
 \textbf{Weihan Peng\textsuperscript{\dag}},
 \textbf{Yuling Shi\textsuperscript{\dag}},
 \textbf{Yuhang Wang},\\
 \textbf{Xinyun Zhang},
 \textbf{Beijun Shen},
 \textbf{Xiaodong Gu\textsuperscript{{\textrm{\Letter}}}}
\\
 Shanghai Jiao Tong University
\\
 \small{
   \texttt{\{peng-weihan, yuling.shi, lingbo\_2022, xinyunz, bjshen, xiaodong.gu\}@sjtu.edu.cn}
 }
 \\
  \faGithub\ GitHub: \href{https://github.com/peng-weihan/SWE-QA-Bench}{https://github.com/peng-weihan/SWE-QA-Bench}
  \\
   \hficon\ Hugging Face: \href{https://huggingface.co/datasets/swe-qa/SWE-QA-Benchmark}{https://huggingface.co/datasets/swe-qa/SWE-QA-Benchmark}
}

\begin{document}

\newcommand{\ourbench}[0]{\textsc{SWE-QA}\xspace}
\newcommand{\todoc}[2]{{\textcolor{#1}{\textbf{[#2]}}}}
\newcommand{\todo}[1]{\todoblue{#1}} 
\newcommand{\todoblue}[1]{\todoc{blue}{\textbf{#1}}}
\newcommand{\todored}[1]{\todoc{red}{\textbf{#1}}}
\newcommand{\todoorange}[1]{\todoc{orange}{\textbf{#1}}}

\newcommand{\peng}[1]{\todoblue{peng: #1}}
\newcommand{\gu}[1]{\todored{gu: #1}}
\newcommand{\shen}[1]{\todored{shen: #1}}
\newcommand{\shi}[1]{\todoorange{shi: #1}}




\maketitle
{\renewcommand{\thefootnote}{}
\footnotetext{\textsuperscript{\dag}Equal contribution. {\textrm{\Letter}}\,Corresponding author.}}
\begin{abstract}
Understanding and reasoning about entire software repositories is an essential capability for intelligent software engineering tools. While existing benchmarks such as CoSQA and CodeQA have advanced the field, they predominantly focus on small self-contained code snippets. These setups fail to capture the complexity of real-world repositories, where effective understanding and reasoning often require navigating multiple files, understanding software architecture, and grounding answers in long-range code dependencies.
In this paper, we present \ourbench, a repository-level code question answering (QA) benchmark designed to facilitate research on automated QA systems in realistic code environments. \ourbench involves 720 high-quality question-answer pairs spanning diverse categories, including intention understanding, cross-file reasoning, and multi-hop dependency analysis. To construct \ourbench, we first crawled 77,100 GitHub issues from 12 popular repositories. Based on an analysis of naturally occurring developer questions extracted from these issues, we developed a two-level taxonomy of repository-level questions and constructed a set of seed questions for each category. For each category, we manually curated and validated questions and collected their corresponding answers. We evaluated six advanced LLMs on \ourbench under various context augmentation strategies. Experimental results highlight the promise of LLMs.
\end{abstract}

\begin{figure}
    \centering
    \includegraphics[width=1\linewidth]{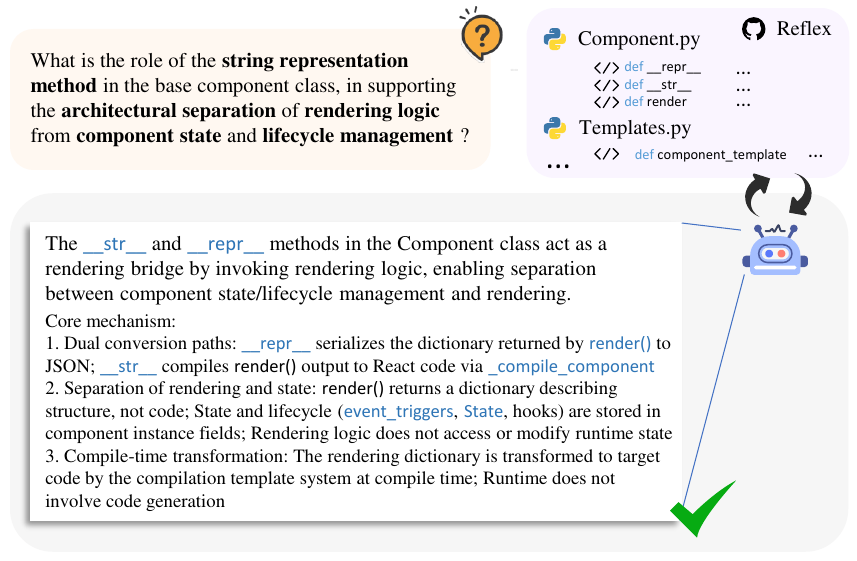}
    \vspace{-0.6cm}
    \caption{Repository-level Code QA Example}
    \vspace{-0.5cm}
    \label{fig:QA-example}
\end{figure}

\begin{figure*}
    \centering
    \includegraphics[width=1\linewidth]{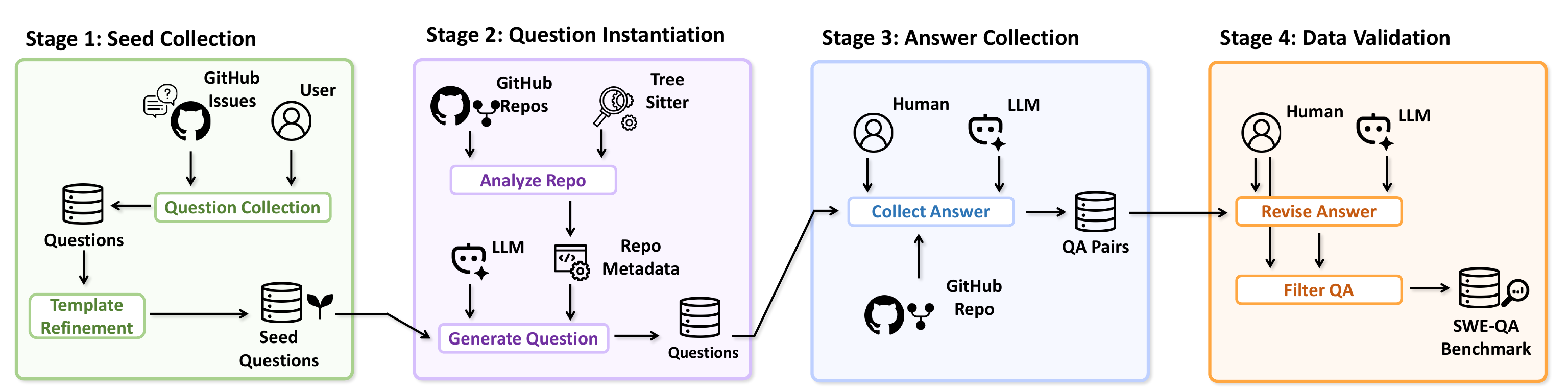}
    \caption{Workflow of benchmark construction.}
    \vspace{-0.5cm}
    \label{fig:dataset-construction}
\end{figure*}

\section{Introduction}\label{sec:intro}


Understanding and reasoning about entire software repositories is an essential capability for intelligent software engineering tools. Real-world development rarely involves reasoning over isolated functions or small code snippets; instead, developers must navigate large, interconnected codebases, trace dependencies across multiple files, and synthesize architectural knowledge to answer complex questions.

Recent advances in large language models (LLMs) have shown promise for code understanding~\cite{feng2020codebert,wang2021codet5,zhu2024deepseek,huang2024opencoder,yang2025elaboration,shi2026codeocr,shi2025longcodezip,zeng2026codesumm}, yet most existing evaluations~\cite{CoSQA,CodeQA,CodeQueries,CS1QA,li2024procqa} target isolated code snippets, functions, or APIs. These benchmarks fail to capture the complexity of real-world repositories, including architecture, cross-file dependencies, lifecycle flows, and design rationales, which require a deeper, multi-hop understanding of code structure, semantics, and intent~\cite{RepoCoder,ouyang2024repograph}. 
While recent works like CodeRepoQA~\cite{hu2024coderepoqa}, CoreQA~\cite{chen2025coreqa}, and Spyder-CodeQA~\cite{strich2024improving} have begun to address repository-level understanding, they still lack comprehensive coverage of the diverse reasoning patterns and multi-hop dependencies essential for realistic software development scenarios. 

To address this limitation, we propose a repository-level code question answering (QA) benchmark designed to evaluate LLMs on realistic repository-based questions. 
SWE-QA comprises 720 high-quality QA pairs that necessitate a deep understanding of intentions, cross-file reasoning, and multi-hop dependency analysis.
To capture the diverse reasoning requirements inherent in real world, we crawled 77,100 GitHub issues from 12 repositories used by SWE-bench~\cite{jimenez2024swe}. Based on an analysis of naturally occurring questions raised from these issues, we developed a two-level taxonomy of repository-level questions and constructed a set of seed questions for each category. 
Guided by our taxonomy and seed templates, we instantiated candidate questions from 15 repositories (12 from SWE-Bench and 3 from SWE-Bench-Live~\cite{zhang2025swe}) using LLMs, then manually screened and refined them to obtain 48 high-quality QA pairs per repository. Each question was answered based on retrieved context by a strong LLM, and preliminary answers were reviewed for correctness, completeness, and clarity. This process produces high-quality reference answers grounded in code context, forming a reliable and scalable benchmark with diverse reasoning requirements.

We evaluated six advanced LLMs on \ourbench using various context augmentation methods. Direct prompting performs poorly, while standard RAG methods significantly improve performance. Agent-based frameworks further enhance results, with OpenHands achieving 70.79 using GPT-5.1. These results are further validated by human evaluation, which shows a high level of agreement with LLM-as-a-Judge scores, indicating the reliability of LLM-based evaluation. Further analysis shows that models perform well on open-ended \textit{How} and \textit{Why} questions but struggle with constrained \textit{What} and locational \textit{How} questions requiring multi-hop reasoning. Performance also varies across repositories, with some like ``pylint'' proving particularly challenging. Overall, the experimental results highlight the promise of LLMs, particularly in the agent framework, in addressing repository-level QA, while also revealing open challenges and pointing to future research directions.


In summary, our contributions are as follows:
\begin{itemize}[leftmargin=20pt, itemsep=0pt, topsep=0pt, parsep=0pt]
    \item \textbf{\ourbench}: a repository-level code QA benchmark comprising 720 high-quality question-answer pairs from 15 diverse open-source Python repositories for evaluating comprehensive repository understanding.
    \item Flexible QA generation pipeline: enables efficient creation of question-answer datasets for any new repository using seed questions.
    \item LLM evaluation under context augmentation: assesses the ability of LLMs to answer questions with various methods, including RAG and diverse Agent Frameworks.
\end{itemize}


\section{Benchmark Construction}
In this section, we introduce \ourbench, a novel benchmark designed for repository-level code question answering. 
As shown in Figure~\ref{fig:dataset-construction}, our benchmark construction pipeline consists of four main stages: seed question collection, question instantiation, answer collection, and data validation. Each of these stages is detailed in the following subsections.

\subsection{Seed Collection and Taxonomy Construction}
\label{sec:stage1}
To ensure \ourbench{} reflects the complexities of real-world software engineering, we first conducted an empirical study to understand the types of questions developers pose when working with large codebases. We systematically collected and analyzed a large corpus of questions from GitHub issues to build a comprehensive taxonomy of repository-level questions. This taxonomy serves as the foundation for our benchmark construction.

Our data collection process began by crawling 77,100 GitHub issues from the 12 popular repositories used in SWE-bench~\cite{jimenez2024swe} (details of collected issues are available in Appendix~\ref{app:collecting-issue}). To focus on substantive discussions, we filtered for issues with a body length of at least 1,000 characters, resulting in a dataset of 41,955 issues. Given that issues often contain extensive descriptive text, we employed a large language model to parse each issue and extract explicit questions related to code understanding (see Prompt~\ref{prompt1} in Appendix~\ref{app:prompt} for details). This process yielded 127,415 distinct questions, with an average of 3.04 questions per issue. 

\begin{figure}
    \centering
    \includegraphics[width=\linewidth]{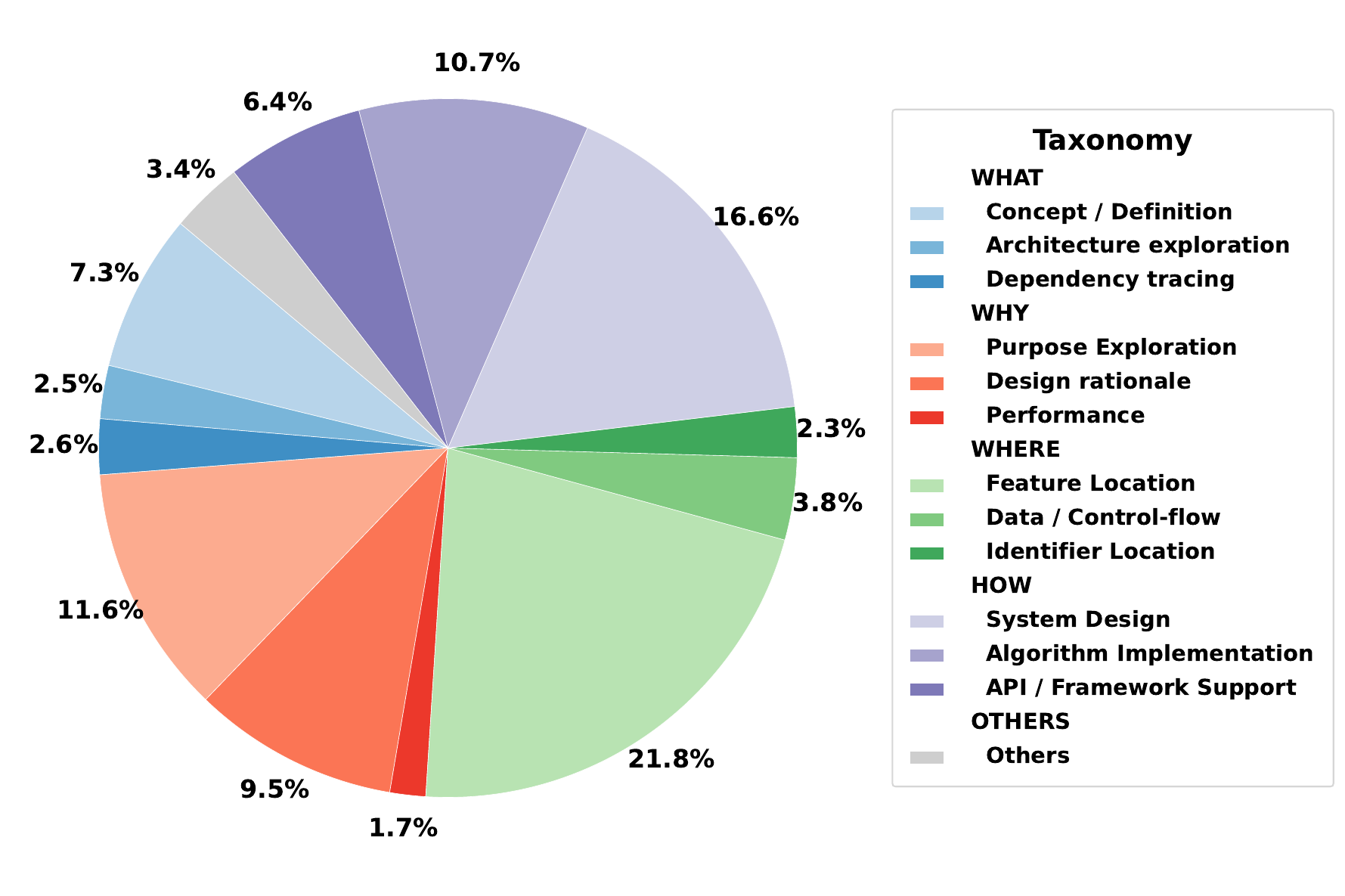}
    \caption{Distribution of question types.}
    \vspace{-0.5cm}
    \label{fig:taxonomy-distribution}
\end{figure}

We first randomly sampled 1,000 questions, which were manually analyzed by two annotators with three years of development experience using an iterative top-down and bottom-up open coding process to identify recurring patterns and developer intentions. The two annotators performed the coding in parallel, and any disagreements were resolved through discussion among the authors until full consensus was reached. This yielded a structured two-level taxonomy for repository-level QA, as summarized in Table~\ref{tab:question_taxonomy}. The first level categorizes questions based on their primary interrogative: \textit{What}, \textit{Why}, \textit{Where}, and \textit{How}. The second level further classifies questions into 12 fine-grained user intentions, such as dependency tracing, design rationale clarification, and algorithm analysis, which reflect common software engineering activities.

\begin{table*}[t]
\centering
\scriptsize 
\resizebox{\textwidth}{!}{%
\renewcommand{\arraystretch}{1.1}
\begin{tabular}{l >{\raggedright\arraybackslash}p{2.5cm} >{\raggedright\arraybackslash}p{3cm} >{\raggedright\arraybackslash}p{9cm}}
\toprule
\textbf{Type} & \textbf{Intention} & \textbf{Definition} & \textbf{Example} \\
\midrule
& \cellcolor{gray!15} Architecture exploration & \cellcolor{gray!15} Identify components or structures of the system & \cellcolor{gray!15} What is the semantic relationship between the expiration threshold computed in the test initialization method and the early-exit optimization tested across multiple test methods? \\
What & Concept / Definition  & Understand meaning of code elements & What does the method that enables the test helper class representing pathlib.Path objects return when invoked by the standard library path conversion function? \\

 & \cellcolor{gray!15} Dependency tracing & \cellcolor{gray!15} Relationships or dependencies among code elements & \cellcolor{gray!15} What is the dependency chain in the test function that verifies subdomain routing through configuration, route registration, and request context creation? \\
\midrule 

 & Design rationale  & Explain why certain design decisions are made & Why does the assignment name processing method defer frame node visitation until the local variable type attribute is accessed rather than visiting frames upfront? \\
 Why & \cellcolor{gray!15} Purpose Exploration & \cellcolor{gray!15} Understand function or module purpose & \cellcolor{gray!15} Why does the guess\_filename function's validation logic prevent security vulnerabilities when extracting filenames from file-like objects in multipart form data handling? \\
 & Performance &Understand performance considerations & Why does the preprocessing transformer that generates univariate B-spline basis functions use an identity coefficient matrix during the fitting process? \\
\midrule 
 & \cellcolor{gray!15} Data / Control-flow & \cellcolor{gray!15} Localize variables or control statements & \cellcolor{gray!15} Where in the serialization pipeline for the JSON tag class that handles Markup API objects are the lower-level helper functions that the conversion method delegates to?\\
 Where & Feature Location & Identify where a feature is implemented& Where in the AxLine class is the coordinate transformation logic applied to reconcile between data coordinates and display coordinates before computing the line endpoints? \\
 & \cellcolor{gray!15} Identifier Location & \cellcolor{gray!15} Find where an identifier is defined or used & \cellcolor{gray!15} Where in the LocalPath class can I locate the code logic responsible for assigning the absolute path value to the strpath attribute during initialization? \\
\midrule 
 & System Design &Explain overall system operation & How does the abstract base class of muscle activation models decouple symbolic equation representation from numerical solver implementations via abstract property interfaces? \\
 How & \cellcolor{gray!15} Algorithm Implementation & \cellcolor{gray!15} Understand algorithm steps & \cellcolor{gray!15} How does the documentation generation system determine which class members should be excluded from generated documentation? \\
 & API / Framework Support & Show usage of APIs or frameworks & How does the test helper function validating dialect-specific SQL statements ensure the parsed tree matches the expected segment hierarchy and statement count via recursive validation? \\
\bottomrule
\end{tabular}
}   
\vspace{-0.25cm}
\caption{Taxonomy of repository-level questions, including intention, definition, and example for each type.}
\vspace{-0.25cm}
\label{tab:question_taxonomy}
\end{table*}

With the taxonomy fixed, we then used a strong LLM (GPT-5) to classify the remaining 126,415 questions into the Level-1/Level-2 categories using concise labeling prompts with consistency checks. The resulting distribution, illustrated in Figure~\ref{fig:taxonomy-distribution}, reveals that \textit{How} questions are the most frequent (35.2\%), focusing on implementation details like system design and algorithms. \textit{Where} questions follow at 28.4\%, indicating that developers often need to locate features, data flows, or specific identifiers. \textit{Why} questions, which probe design rationales and purpose, make up 23.1\% of the corpus. Finally, \textit{What} questions, seeking definitions or architectural summaries, account for the remaining 13.3\%. This distribution underscores that a significant portion of developer queries are centered on procedural and locational knowledge, highlighting the need for QA systems that can reason deeply about code structure and implementation.

Based on this taxonomy, we created a set of abstract seed templates for each user intention category. These templates, such as ``What are the subclasses that inherit from the <Class> class?'' and ``How does <Module> implement <Feature> in case of <Condition>?'', capture the essence of recurring questions. As detailed in Table~\ref{tab:question_taxonomy}, these templates serve as the blueprint for generating diverse, context-specific question instances tailored to individual repositories.

\subsection{Question Instantiation and Expansion}

The objective of this stage is to generate context-specific question instances tailored to a target repository. To extract relevant contextual information, we parse the structure of each repository using \texttt{tree-sitter}\footnote{\url{https://tree-sitter.github.io/tree-sitter/}}(see Appendix~\ref{app:repo-structure} for details). We instantiate questions by selecting a compact subgraph around a focal element (e.g., a class or method) and combining it with seed templates from Stage 1. The subgraph ensures sufficient context without overwhelming the prompt (see Prompt~\ref{prompt2} in Appendix~\ref{app:prompt} for details). For a concrete example of question generation, we refer to Appendix~\ref{app:question_example}, where a full instance is presented.





\subsection{Answer Collection}

Once the questions are instantiated, we proceed to generate initial reference answers for each question. This stage, illustrated as Stage 3 in Figure~\ref{fig:dataset-construction}, leverages a retrieval-augmented generation (RAG) pipeline designed to ground answers firmly in the repository's context. The process involves two key steps:

\textit{\textbf{Step 1: Context Retrieval}}. For each question, we first build a comprehensive index of the target repository's code elements, including functions, classes, and their inter-dependencies. Using this index, we retrieve relevant code snippets, documentation, and architectural metadata through a combination of semantic similarity matching and structural dependency analysis. This ensures a rich, relevant context for answer generation.

\textit{\textbf{Step 2: Initial Answer Generation}}. With the retrieved context, we utilize a powerful LLM, assisted by 4 human experts with at least three years of experience in software development using tools like Cursor, to generate a preliminary answer. This process is guided by a prompt that emphasizes factual accuracy, completeness, and strict adherence to the provided context. The prompt explicitly directs the model to cite code locations and avoid introducing information not present in the retrieved materials, thus minimizing hallucination. The resulting question-answer pairs serve as the input for the subsequent data validation stage.

\begin{table*}[t]
    \centering
    \small
    \resizebox{\linewidth}{!}{
    \begin{tabular}{lccccccccc}
        \toprule
        \bf Dataset & \bf Year & \bf Source & \bf \makecell{Test Data\\Size} & \bf \makecell{Repo\\Level?} & \bf \makecell{Module\\Reasoning?} & \bf \makecell{Multi\\-hop?} & \bf \makecell{Cross\\file?} & \bf \makecell{Human\\Verified?} \\
        \midrule
        CoSQA~\cite{CoSQA} & 2021 & Bing Search Logs & 1046 & \ding{55} & \ding{55} & \ding{55} & \ding{55} & $\checkmark$ \\
        CodeQA~\cite{CodeQA} & 2021 & GitHub Code Comments & \makecell{Java: 11,978\\Python: 7,009} & \ding{55} & \ding{55} & \ding{55} & \ding{55} & $\checkmark$ \\
        CodeQueries~\cite{CodeQueries} & 2022 & \makecell{ETH Py150 Open\\(GitHub Python code)} & 29,033 & \ding{55} & \ding{55} & $\checkmark$ & \ding{55} & \ding{55} \\
        CS1QA~\cite{CS1QA} & 2022 & \makecell{Python Programming\\Courses Chat Logs} & 1,847 & \ding{55} & \ding{55} & \ding{55} & \ding{55} & $\checkmark$ \\
        ProCQA~\cite{li2024procqa} & 2024 & StackOverflow & $\sim$500,000 & \ding{55} & \ding{55} & \ding{55} & \ding{55} & $\checkmark$ \\
        InfiBench~\cite{li2024infibench} & 2024 & Stack Overflow & 234 & \ding{55} & \ding{55} & \ding{55} & \ding{55} & $\checkmark$ \\
        CoReQA~\cite{chen2025coreqa} & 2025 & \makecell{GitHub Issues\\and Comments} & 1,563 & $\checkmark$ & \ding{55} & \ding{55} & $\checkmark$ & $\checkmark$ \\
        \ourbench (Ours) & 2025 & \makecell{GitHub Repositories} & 720 & $\checkmark$ & $\checkmark$ & $\checkmark$ & $\checkmark$ & $\checkmark$ \\
        \bottomrule
    \end{tabular}
    }
    \vspace{-0.1cm}
    \caption{Comparison between \ourbench and existing code QA benchmarks. \ourbench focuses on repository-level multi-hop questions that require complex reasoning and deep repository understanding.}
    \vspace{-0.25cm}
    \label{tab:comparison}
\end{table*}
\subsection{Data Validation}
To ensure the highest quality and reliability of our benchmark, all preliminary QA pairs undergo a rigorous data validation process, as depicted in Stage 4 of Figure~\ref{fig:dataset-construction}. This multi-phase procedure is conducted by experienced developers with deep familiarity with the target repositories and involves both answer revision and quality filtering.

\textit{\textbf{Step 1: Expert Answer Revision}}. Each generated answer is manually reviewed by our expert team. With the assistance of LLM-powered tools like Cursor, reviewers meticulously verify the factual accuracy of every claim, assess the completeness of the explanation, and refine the language for clarity and precision. Each question is answered independently by two experts, and their answers are cross‑validated; if there is significant disagreement, a third expert also participates in answer generation and the three discuss to reach a final consensus. This human-in-the-loop approach allows for nuanced corrections that automated systems might miss, ensuring each answer is not only correct and complete but also easy to understand.

\textit{\textbf{Step 2: Quality Filtering}}. After revision, the QA pairs are subjected to a final filtering step. Pairs are discarded if they fail to meet our quality standards. The criteria for filtering include, but are not limited to: questions that are ambiguous or poorly formulated, answers that remain factually incorrect or incomplete after revision, or answers that cannot be sufficiently grounded in the repository's code and documentation. We also enforce per-repository balance across Level-1 categories (\textit{What}, \textit{Why}, \textit{Where}, \textit{How}), yielding exactly 48 finalized pairs per repository. This stringent filtering ensures that only the most clear, correct, and valuable QA pairs are included in the final \ourbench{} benchmark.


\subsection{Statistics of \ourbench}
The benchmark comprises 720 questions meticulously curated from 15 Python repositories. In total, these repositories encompass 13,300 files, 22,522 classes, 142,404 functions, and over 3.4 million lines of code, presenting a substantial and realistic challenge for code understanding models. The average question length is 28.62 words, and the average answer length is 266.64 words. To further characterize the complexity of \ourbench, we conduct a quantitative analysis of the reasoning chain and structural requirements of its questions. On average, answering each question involves 8.71 functions across 3.19 files, with a reasoning chain depth of 4.72 and a dependency chain depth of 2.96. Moreover, 90.9\% of the questions exhibit a reasoning chain depth greater than one, and 77.6\% require cross-file knowledge to be answered correctly, indicating that \ourbench systematically emphasizes multi-hop reasoning and non-trivial code dependencies. To ensure a balanced representation of question types, we selected an equal number of What, Why, Where, and How questions, with each repository contributing exactly 48 samples. Detailed statistics for each repository can be found in Appendix~\ref{app:bench-static}.

Table~\ref{tab:comparison} presents a comprehensive comparison between \ourbench and existing code QA benchmarks. While traditional benchmarks like CoSQA~\cite{CoSQA} and CodeQA~\cite{CodeQA} focus on isolated code snippets, \ourbench operates at the repository level, similar to recent benchmarks such as CodeRepoQA~\cite{hu2024coderepoqa} and CoreQA~\cite{chen2025coreqa}. However, \ourbench is unique in its comprehensive design, incorporating categorized, multi-hop questions that require cross-file context, with all question-answer pairs being human-verified. This multi-faceted design addresses the limitations of existing repository-level benchmarks, which often lack one or more of these critical dimensions. For instance, CodeRepoQA~\cite{hu2024coderepoqa} lacks human verification and categorization, while Spyder-CodeQA~\cite{strich2024improving} and CoreQA~\cite{chen2025coreqa} do not feature multi-hop questions.
\ourbench's realistic construction setting imbues the dataset with the following unique features:

\textit{\textbf{Repository-Level Granularity and Complexity.}}
Unlike benchmarks targeting isolated code elements (e.g., functions or classes), \ourbench evaluates repositories, reflecting real-world tasks that require cross-file context, architectural understanding, and dependency tracking. This design challenges models to answer questions in complex, interconnected codebases, with substantially higher cognitive and structural demands than existing code QA datasets.

\textit{\textbf{Pipeline-Level Extensibility.}}
\ourbench serves as both a benchmark and a modular pipeline for generating new repository-level QA instances applicable to any open-source project. Leveraging static code analysis, LLM prompting, and human filtering, new benchmarks can be continuously and semi-automatically created, ensuring scalability and adaptability to evolving codebases.

\section{Evaluation}

To showcase the usefulness of \ourbench, we assess the performance of language models on repository-level code question answering using the proposed benchmark. Our objective is to uncover novel insights that have not been previously explored through comprehensive and in-depth comparisons of existing language models.


\subsection{Experiment Setup}

\paragraph{Model Selection.}
We evaluate six representative LLMs: Kimi K2~\cite{kimi-k2-0905-preview}, Qwen3-Coder-30B-A3B-Instruct~\cite{qwen3-coder-30b-a3b-instruct}, Qwen3-Coder-480B-A35B-Instruct~\cite{qwen3-coder-480b-a35b-instruct}, GLM-4.6~\cite{glm-4.6}, Gemini 2.5 Pro~\cite{gemini-2.5-pro} and GPT-5.1~\cite{gpt-5.1-2025-11-13}.

\paragraph{Studied Methods.}
For each model, we study five approaches: \textit{Direct Prompting}, two retrieval-augmented generation variants (\textit{Sliding Window RAG} and \textit{Function Chunking RAG}), and two agent-based frameworks (\textit{OpenHands} and \textit{SWE-agent}). 
We further include two commercial programming tools, Tongyi Lingma\footnote{\url{https://lingma.aliyun.com/}, v2.5.16} and Cursor-agent\footnote{\url{https://cursor.com}, v2025.09.04}, as system-level baselines. 
Detailed configurations are reported in Appendix~\ref{app:experimental-details}.

\paragraph{Metrics.}
\label{sec:dimensions}
We adopt an LLM-based evaluation (LLM-as-Judge) to assess repository-level code question answering performance (see Prompt~\ref{prompt3}). Prior studies have demonstrated the reliability of LLM-as-Judge across both natural language generation~\cite{liu2023g, song2024finesure, dai2025psycher1, lan2025f2bench} and software engineering tasks~\cite{wang2025can, he2025code}. Given a model’s output and the gold answer, an LLM—Claude Sonnet 4.5~\cite{claudeSonnet45} in our experiments—evaluates answer quality along five dimensions: \textbf{correctness}, \textbf{completeness}, \textbf{relevance}, \textbf{clarity}, and \textbf{Coherence}. Each dimension is scored on a 20-point scale, yielding a total score of 100 (see Appendix~\ref{app:llm-as-judge} for details). To mitigate potential self-evaluation bias, we enforce strict judge–candidate separation, anonymize systems, randomize answer order, and complement automated evaluation with a human study (Appendix~\ref{sec:human_evaluation}).

\begin{table*}[t]
    \centering
    \small
    \renewcommand{\arraystretch}{1.05}
    \resizebox{\linewidth}{!}{
    \begin{tabular}{l@{\hspace{3mm}}c@{\hspace{3mm}}c@{\hspace{3mm}}c@{\hspace{3mm}}c@{\hspace{3mm}}c@{\hspace{3mm}}c}
        \toprule
        \multirow{2}[2]{*}{\textbf{Model}} & \multicolumn{5}{c}{\textbf{Evaluation Metrics}} & \multirow{2}[2]{*}{\hspace{2mm} \textbf{Overall}} \\
        \cmidrule{2-6} 
        & \textbf{Correctness} & \textbf{Completeness} & \textbf{Relevance} & \textbf{Clarity} & \textbf{Reasoning} & \\ 
        \midrule
        \multicolumn{7}{l}{\textit{Commercial Tools}} \\
        \midrule
        Tongyi Lingma & 12.02 & 10.99 & 16.78 & 15.67 & 13.61 & 69.07 \\
        Cursor & 12.11 & 11.44 & 17.21 & 16.01 & 13.89 & 70.66 \\
        \midrule
        \multicolumn{7}{l}{\textit{Open-Source Frameworks}} \\
        \midrule

        Qwen3-Coder-30B-A3B-Instruct & 7.39 & 5.42 & 14.23 & 14.66 & 9.11 & 50.80 \\
        \hspace{1em} + Function Chunking RAG & 10.88 (\textcolor{green!50!black}{+3.49}) & \, 8.72 (\textcolor{green!50!black}{+3.30}) & 15.61 (\textcolor{green!50!black}{+1.38}) & 16.24 (\textcolor{green!50!black}{+1.58}) & 11.95 (\textcolor{green!50!black}{+2.84}) & 63.40 (\textcolor{green!50!black}{+12.60}) \\
        \hspace{1em} + Sliding Window RAG & 11.22 (\textcolor{green!50!black}{+3.83}) & \, 9.00 (\textcolor{green!50!black}{+3.58}) & 15.81 (\textcolor{green!50!black}{+1.58}) & 16.71 (\textcolor{green!50!black}{+2.05}) & 12.12 (\textcolor{green!50!black}{+3.01}) & 64.86 (\textcolor{green!50!black}{+14.06}) \\
        \hspace{1em} + SWE-agent & \, 9.12 (\textcolor{green!50!black}{+1.73}) & \, 8.64 (\textcolor{green!50!black}{+3.22}) & 14.98 (\textcolor{green!50!black}{+0.75}) & 12.92 (\textcolor{red}{-1.74}) & 11.77 (\textcolor{green!50!black}{+2.66}) & 57.44 (\textcolor{green!50!black}{+\, 6.64}) \\
        \hspace{1em} + OpenHands & 11.09 (\textcolor{green!50!black}{+3.70}) & 10.40 (\textcolor{green!50!black}{+4.98}) & 15.94 (\textcolor{green!50!black}{+1.71}) & 15.08 (\textcolor{red}{-0.58}) & 13.36 (\textcolor{green!50!black}{+4.25}) & 65.88 (\textcolor{green!50!black}{+15.08}) \\
        \midrule

        Qwen3-Coder-480B-A35B-Instruct & 8.14 & 6.09 & 15.00 & 15.50 & 9.88 & 54.61 \\
        \hspace{1em} + Function Chunking RAG & 10.61 (\textcolor{green!50!black}{+2.47}) & \, 8.56 (\textcolor{green!50!black}{+2.47}) & 16.08 (\textcolor{green!50!black}{+1.08}) & 16.07 (\textcolor{green!50!black}{+0.57}) & 11.63 (\textcolor{green!50!black}{+1.75}) & 62.96 (\textcolor{green!50!black}{+\, 8.35}) \\
        \hspace{1em} + Sliding Window RAG & 11.36 (\textcolor{green!50!black}{+3.22}) & \, 9.11 (\textcolor{green!50!black}{+3.02}) & 16.34 (\textcolor{green!50!black}{+1.34}) & 16.51 (\textcolor{green!50!black}{+1.01}) & 12.06 (\textcolor{green!50!black}{+2.18}) & 65.39 (\textcolor{green!50!black}{+10.78}) \\
        \hspace{1em} + SWE-agent & 11.09 (\textcolor{green!50!black}{+2.95}) & 10.59 (\textcolor{green!50!black}{+4.50}) & 15.49 (\textcolor{green!50!black}{+0.49}) & 13.76 (\textcolor{red}{-1.74}) & 13.06 (\textcolor{green!50!black}{+3.18}) & 64.00 (\textcolor{green!50!black}{+\, 9.39}) \\
        \hspace{1em} + OpenHands & 11.89 (\textcolor{green!50!black}{+3.75}) & 11.20 (\textcolor{green!50!black}{+5.11}) & 16.21 (\textcolor{green!50!black}{+1.21}) & 15.37 (\textcolor{red}{-0.13}) & 13.90 (\textcolor{green!50!black}{+4.02}) & 68.57 (\textcolor{green!50!black}{+13.96}) \\
        \midrule

        Kimi K2 & 7.41 & 4.93 & 15.23 & 15.75 & 8.15 & 51.47 \\
        \hspace{1em} + Function Chunking RAG & 9.78 (\textcolor{green!50!black}{+2.37}) & \, 7.93 (\textcolor{green!50!black}{+3.00}) & 16.55 (\textcolor{green!50!black}{+1.32}) & 15.11 (\textcolor{red}{-0.64}) & 10.70 (\textcolor{green!50!black}{+2.55}) & 60.08 (\textcolor{green!50!black}{+\, 8.61}) \\
        \hspace{1em} + Sliding Window RAG & 10.44 (\textcolor{green!50!black}{+3.03}) & \, 8.55 (\textcolor{green!50!black}{+3.62}) & 16.48 (\textcolor{green!50!black}{+1.25}) & 15.65 (\textcolor{green!50!black}{-0.10}) & 11.32 (\textcolor{green!50!black}{+3.17}) & 62.44 (\textcolor{green!50!black}{+10.97}) \\
        \hspace{1em} + SWE-agent & 11.84 (\textcolor{green!50!black}{+4.43}) & 11.72 (\textcolor{green!50!black}{+6.79}) & 15.88 (\textcolor{green!50!black}{+0.65}) & 14.48 (\textcolor{red}{-1.27}) & 13.80 (\textcolor{green!50!black}{+5.65}) & 67.72 (\textcolor{green!50!black}{+16.25}) \\
        \hspace{1em} + OpenHands & 11.74 (\textcolor{green!50!black}{+4.33}) & 11.33 (\textcolor{green!50!black}{+6.40}) & 15.21 (\textcolor{red}{-0.02}) & 14.66 (\textcolor{red}{-1.09}) & 13.24 (\textcolor{green!50!black}{+5.09}) & 66.18 (\textcolor{green!50!black}{+14.71}) \\
        \midrule

        GLM-4.6 & 8.71 & 6.09 & 15.79 & 16.53 & 9.53 & 56.64 \\
        \hspace{1em} + Function Chunking RAG & \, 9.48 (\textcolor{green!50!black}{+0.77}) & \, 7.52 (\textcolor{green!50!black}{+1.43}) & 16.07 (\textcolor{green!50!black}{+0.28}) & 14.41 (\textcolor{red}{-2.12}) & \, 9.99 (\textcolor{green!50!black}{+0.46}) & 57.46 (\textcolor{green!50!black}{+\, 0.82}) \\
        \hspace{1em} + Sliding Window RAG & \, 9.02 (\textcolor{green!50!black}{+0.31}) & \, 7.85 (\textcolor{green!50!black}{+1.76}) & 16.33 (\textcolor{green!50!black}{+0.54}) & 16.08 (\textcolor{red}{-0.45}) & 10.44 (\textcolor{green!50!black}{+1.91}) & 60.72 (\textcolor{green!50!black}{+\, 3.08}) \\
        \hspace{1em} + SWE-agent & 12.31 (\textcolor{green!50!black}{+3.60}) & 12.37 (\textcolor{green!50!black}{+6.28}) & 16.15 (\textcolor{green!50!black}{+0.36}) & 14.81 (\textcolor{red}{-1.72}) & 14.21 (\textcolor{green!50!black}{+4.68}) & 69.85 (\textcolor{green!50!black}{+13.21}) \\
        \hspace{1em} + OpenHands & 11.91 (\textcolor{green!50!black}{+3.20}) & 12.70 (\textcolor{green!50!black}{+6.61}) & 16.90 (\textcolor{green!50!black}{+1.11}) & 13.82 (\textcolor{red}{-2.71}) & 14.82 (\textcolor{green!50!black}{+5.29}) & 70.15 (\textcolor{green!50!black}{+13.51}) \\
        \midrule

        Gemini 2.5 Pro & 7.84 & 5.59 & 16.28 & 15.83 & 9.18 & 54.71 \\
        \hspace{1em} + Function Chunking RAG & \, 9.91 (\textcolor{green!50!black}{+2.07}) & \, 7.53 (\textcolor{green!50!black}{+1.94}) & 16.79 (\textcolor{green!50!black}{+0.51}) & 15.32 (\textcolor{red}{-0.51}) & 10.47 (\textcolor{green!50!black}{+1.29}) & 60.02 (\textcolor{green!50!black}{+\, 5.31}) \\
        \hspace{1em} + Sliding Window RAG & 11.52 (\textcolor{green!50!black}{+3.68}) & \, 8.83 (\textcolor{green!50!black}{+3.24}) & 16.85 (\textcolor{green!50!black}{+0.57}) & 16.66 (\textcolor{green!50!black}{+0.83}) & 11.82 (\textcolor{green!50!black}{+2.64}) & 65.68 (\textcolor{green!50!black}{+10.97}) \\
        \hspace{1em} + SWE-agent & 10.62 (\textcolor{green!50!black}{+2.78}) & 10.11 (\textcolor{green!50!black}{+4.52}) & 15.96 (\textcolor{red}{-0.32}) & 13.81 (\textcolor{red}{-2.02}) & 13.26 (\textcolor{green!50!black}{+4.08}) & 63.76 (\textcolor{green!50!black}{+\, 9.05}) \\
        \hspace{1em} + OpenHands & 10.75 (\textcolor{green!50!black}{+2.91}) & \, 9.64 (\textcolor{green!50!black}{+4.05}) & 16.78 (\textcolor{green!50!black}{+0.50}) & 15.20 (\textcolor{red}{-0.63}) & 13.27 (\textcolor{green!50!black}{+4.09}) & 65.63 (\textcolor{green!50!black}{+10.92}) \\
        \midrule

        GPT-5.1 & 9.62 & 7.37 & 16.23 & 16.86 & 11.33 & 61.41 \\
        \hspace{1em} + Function Chunking RAG & 11.38 (\textcolor{green!50!black}{+1.76}) & \, 9.50 (\textcolor{green!50!black}{+2.13}) & 17.12 (\textcolor{green!50!black}{+0.89}) & 16.15 (\textcolor{red}{-0.71}) & 12.43 (\textcolor{green!50!black}{+1.10}) & 66.57 (\textcolor{green!50!black}{+5.16}) \\
        \hspace{1em} + Sliding Window RAG & 11.45 (\textcolor{green!50!black}{+2.38}) & \, 9.44 (\textcolor{green!50!black}{+2.07}) & 17.01 (\textcolor{green!50!black}{+0.78}) & 16.28 (\textcolor{red}{-0.58}) & 12.61 (\textcolor{green!50!black}{+1.28}) & 66.79 (\textcolor{green!50!black}{+5.38}) \\
        \hspace{1em} + SWE-agent & 12.94 (\textcolor{green!50!black}{+3.32}) & 11.44 (\textcolor{green!50!black}{+4.07}) & 14.80 (\textcolor{red}{-1.43}) & 16.34 (\textcolor{red}{-0.52}) & 13.68 (\textcolor{green!50!black}{+2.35}) & 69.20 (\textcolor{green!50!black}{+7.79}) \\
        \hspace{1em} + OpenHands & 12.26 (\textcolor{green!50!black}{+2.64}) & 11.60 (\textcolor{green!50!black}{+4.23}) & 16.82 (\textcolor{green!50!black}{+0.59}) & 15.59 (\textcolor{red}{-1.27}) & 14.52 (\textcolor{green!50!black}{+3.19}) & 70.79 (\textcolor{green!50!black}{+9.38}) \\
        \midrule
        
    \end{tabular}
    }
    \caption{Comparative performance of different language models on \ourbench. GPT-5.1 achieves the best performance among all models. In general, RAG-enhanced methods outperform direct inference, and agent-based methods further improve upon RAG methods.}
    \vspace{-0.1cm}
    \label{tab:rq1_main}
\end{table*}

\subsection{Performance of Language Models}
Table~\ref{tab:rq1_main} summarizes the overall performance of different language models on all QA pairs in \ourbench, revealing clear performance gaps across both methods and model choices for repository-level question answering.

\paragraph{Impact of Context Augmentation Methods.} 
We first examine the effect of context augmentation. Direct prompting without repository context consistently yields the lowest performance across all models (e.g., Kimi K2 scores 51.47/100), highlighting the necessity of grounded code context. Both Sliding Window RAG and Function Chunking RAG substantially improve results by retrieving relevant code snippets (e.g., Kimi K2 improves to 62.44 with Sliding Window RAG). Agent-based frameworks further boost performance by enabling iterative reasoning and tool use, leading to the best results among open methods (e.g., Kimi K2 reaches 67.72 with SWE-agent). Overall, richer and more structured context access translates into steadily improved answer quality.

\paragraph{Impact of Language Model Choice.} 
Model choice also plays a critical role. GPT-5.1 achieves the best overall performance, reaching 70.79 when combined with OpenHands. Notably, strong open-source models perform competitively: GLM-4.6 with OpenHands attains 70.15, closely matching GPT-5.1. In contrast, smaller models such as Qwen3-Coder-30B-A3B-Instruct benefit less from agent frameworks, with agent-based methods performing on par with or worse than RAG. This suggests that effective agent planning, tool invocation, and long-term memory remain challenging for smaller-capacity models.

\paragraph{Performance of Commercial Programming Tools.}
We further evaluate two commercial programming tools, Tongyi Lingma and Cursor. Both exhibit strong performance, with Cursor achieving 70.66—second only to GPT-5.1 with OpenHands—and Tongyi Lingma scoring 69.07. As highly integrated systems combining proprietary LLMs with advanced retrieval and orchestration, their results provide a strong reference point and underscore the effectiveness of end-to-end, tool-augmented solutions for complex repository-level question answering.


\begin{table*}[t]
\centering
\small
\resizebox{1\linewidth}{!}{
\begin{tabular}{l|c c c c c c |c}
\toprule
\textbf{Question Type} &
\makecell{\textbf{Qwen3-Coder-}\\\textbf{30B-A3B-Instruct}} &
\makecell{\textbf{Qwen3-Coder-}\\\textbf{480B-A35B-Instruct}} &
\makecell{\textbf{Kimi K2}} &
\makecell{\textbf{GLM-4.6}} &
\makecell{\textbf{Gemini 2.5 Pro}} &
\makecell{\textbf{GPT-5.1}} &
\textbf{Average} \\
\midrule\rowcolor{gray!15}

What
& 62.59 & 67.28 & 64.58 & 67.51 & 64.52 & 68.39 & 65.81 \\ \hline
~Architecture exploration
& 58.95 & 64.24 & 60.25 & 63.40 & 60.33 & 63.89 & 61.84 \\
~Concept / Definition
& 65.60 & 69.94 & 68.30 & 70.86 & 67.47 & 71.10 & 68.88 \\
~Dependency tracing
& 63.19 & 67.72 & 65.14 & 68.34 & 64.84 & 70.17 & 66.57 \\
\hline\rowcolor{gray!15}

Why
& 69.33 & 69.42 & 66.16 & 72.57 & 68.04 & 73.09 & 69.77 \\ \hline
~Design rationale
& 70.90 & 70.72 & 67.54 & 74.75 & 69.72 & 75.26 & 71.48 \\
~Purpose Exploration
& 69.74 & 70.14 & 66.77 & 73.39 & 69.16 & 74.15 & 70.56 \\
~Performance
& 67.33 & 67.43 & 64.06 & 69.50 & 65.21 & 69.90 & 67.24 \\
\hline\rowcolor{gray!15}

Where
& 65.92 & 69.08 & 63.98 & 69.01 & 63.29 & 69.28 & 66.76 \\ \hline
~Data / Control-flow
& 64.44 & 67.82 & 62.10 & 67.18 & 61.83 & 66.94 & 65.05 \\
~Feature Location
& 65.60 & 68.88 & 63.51 & 68.55 & 63.05 & 68.51 & 66.35 \\
~Identifier Location
& 67.72 & 70.62 & 65.25 & 71.28 & 64.93 & 72.39 & 68.70 \\
\hline\rowcolor{gray!15}

How
& 65.68 & 68.50 & 70.00 & 71.51 & 66.68 & 72.40 & 69.13 \\ \hline
~System Design
& 65.31 & 68.01 & 69.49 & 70.76 & 66.15 & 71.75 & 68.58 \\
~Algorithm Implementation
& 66.37 & 68.88 & 70.47 & 72.34 & 67.28 & 73.23 & 69.76 \\
~API / Framework Support
& 65.41 & 68.59 & 70.04 & 71.39 & 65.78 & 71.28 & 68.75 \\

\bottomrule
\end{tabular}
}
\vspace{-0.1cm}
\caption{Results across different question types by OpenHands. What and Where question types present greater challenges. Among the subtypes, Architecture exploration and Data / Control-flow are the most challenging.}
\label{tab:rq3_main}
\end{table*}

\begin{table*}[t]
\small
\centering
\resizebox{\linewidth}{!}{
\begin{tabular}{l|c c c c c c |c}
\toprule
\textbf{Repository} &
\makecell{\textbf{Qwen3-Coder-}\\\textbf{30B-A3B-Instruct}} &
\makecell{\textbf{Qwen3-Coder-}\\\textbf{480B-A35B-Instruct}} &
\makecell{\textbf{Kimi K2}} &
\makecell{\textbf{GLM-4.6}} &
\makecell{\textbf{Gemini 2.5 Pro}} &
\makecell{\textbf{GPT-5.1}} &
\textbf{Average} \\ 
\midrule\rowcolor{gray!15}

From SWE-Bench 
& 66.57 & 69.92 & 66.65 & 70.78 & 66.06 & 71.54 & 68.59 \\ \hline
\hspace{1em} astropy 
& 67.42 & 67.20 & 68.54 & 69.39 & 67.22 & 69.95 & 68.29 \\
\hspace{1em} django 
& 66.92 & 68.02 & 66.67 & 69.54 & 65.02 & 68.15 & 67.39 \\
\hspace{1em} flask 
& 74.38 & 73.50 & 75.41 & 76.99 & 75.34 & 76.91 & 75.42 \\
\hspace{1em} matplotlib 
& 72.13 & 74.07 & 73.99 & 72.26 & 68.45 & 75.81 & 72.78 \\
\hspace{1em} pylint 
& 55.95 & 63.77 & 62.51 & 62.38 & 62.35 & 65.13 & 62.01 \\
\hspace{1em} pytest 
& 60.57 & 71.66 & 59.02 & 71.03 & 57.91 & 71.48 & 65.28 \\
\hspace{1em} requests 
& 70.42 & 76.40 & 68.53 & 78.32 & 67.00 & 79.00 & 73.28 \\
\hspace{1em} scikit-learn 
& 70.08 & 71.65 & 70.98 & 72.60 & 70.58 & 72.17 & 71.34 \\
\hspace{1em} sphinx 
& 64.51 & 65.18 & 62.82 & 66.22 & 65.09 & 69.15 & 65.50 \\
\hspace{1em} sqlfluff 
& 63.53 & 66.58 & 61.87 & 69.70 & 63.65 & 70.62 & 65.99 \\
\hspace{1em} sympy 
& 63.53 & 67.89 & 61.87 & 67.21 & 63.17 & 67.50 & 65.20 \\
\hspace{1em} xarray 
& 69.44 & 73.09 & 67.58 & 73.71 & 67.00 & 72.67 & 70.58 \\ 
\midrule
\rowcolor{gray!15}
From SWE-Bench-Live 
& 63.11 & 63.18 & 64.30 & 67.63 & 63.89 & 67.78 & 64.98 \\ \hline
\hspace{1em} conan 
& 62.28 & 62.55 & 63.98 & 67.49 & 64.03 & 66.72 & 64.51 \\
\hspace{1em} reflex 
& 66.10 & 64.12 & 64.93 & 67.12 & 64.77 & 68.16 & 65.87 \\
\hspace{1em} streamlink 
& 60.93 & 62.87 & 63.99 & 68.28 & 62.88 & 68.45 & 64.57 \\ 
\bottomrule
\end{tabular}
}
\vspace{-0.1cm}
\caption{Results across different repositories by OpenHands. Performance varies substantially across different repositories. Repositories from SWE-Bench-Live present greater challenges compared to those from SWE-Bench.} 
\vspace{-0.1cm}
\label{tab:rq4_main}
\end{table*}





\subsection{Taxonomy-Aware Analysis} 

To understand how performance varies across different types of repository-level questions, we conduct a taxonomy-aware analysis using OpenHands. Table~\ref{tab:rq3_main} breaks down the scores for each model across the 12 question intentions defined in our taxonomy.

The results reveal a discrepancy between high-level, conceptual questions and low-level, implementation-focused queries. Models consistently achieve the highest scores on \textit{Why} questions (average 69.77), particularly ``Design rationale'' (71.48), and perform well on \textit{How} questions related to ``API / Framework Support'' (68.75). This suggests models excel when the required information is explicitly expressed in natural language (e.g., docstrings, comments, architectural notes).

Performance is lower on questions that require deep procedural or locational understanding. \textit{What} questions (e.g., explore architectures or trace dependencies) yield the lowest average score (65.81), with ``Architecture exploration'' at 61.84. \textit{Where} questions (66.76), which demand precise location, also remain challenging. These categories often require reconstructing dispersed logic across files and implicit control paths beyond inline documentation, stressing code tracing and dependency reasoning.

\subsection{Cross-Repository Generalization}

To assess the generalization capabilities of the models, we analyze their performance across the 12 different repositories in \ourbench, again using OpenHands. The results, detailed in Table~\ref{tab:rq4_main}, indicate that while performance is generally consistent, it can be significantly influenced by the specific characteristics of each repository.

On average, most repositories present a similar level of difficulty, with scores clustering around the 70-point mark. For instance, ``django'' (67.39), ``astropy'' (68.29) and ``scikit-learn'' (71.34) show comparable results. However, we observe notable outliers. The ``flask'' repository appears to be easier on average (75.42). Conversely, ``pylint '' (62.01) and ``conan'' (64.51) are more challenging. This variance aligns with factors such as codebase size, architectural complexity, plugin or hook systems, API surface clarity, and unconventional patterns that increase reasoning depth.

Furthermore, the 12 repositories selected from SWE-Bench achieve an average score of 68.59, whereas the 3 repositories drawn from SWE-Bench-Live average 64.98, representing a decrease of 3.61 points. This gap may be related to the reduced data leakage in SWE-Bench-Live, which makes the evaluation more challenging.

\section{Related Work}
Several benchmarks have been developed to evaluate code question answering (QA) systems, from snippet-level to emerging repository-level assessments. CodeQueries~\cite{CodeQueries} evaluates single- and multi-hop reasoning over Python code with annotated answer spans. CoSQA~\cite{CoSQA} uses real user queries from Bing labeled by answer correctness. CodeQA~\cite{CodeQA} generates Q\&A pairs for Python and Java methods via templates. CS1QA~\cite{CS1QA} provides 9,237 Q\&A pairs from introductory Python courses, including tasks like question type classification and code line selection. Recent repository-level benchmarks include CodeRepoQA~\cite{hu2024coderepoqa} with 585,687 entries across five languages, CoreQA~\cite{chen2025coreqa} from GitHub issues across 176 repositories, Spyder-CodeQA~\cite{strich2024improving} with 325 Python Q\&A pairs, InfiBench~\cite{li2024infibench} for freeform real-world questions, and ProCQA~\cite{li2024procqa} from StackOverflow. These benchmarks highlight the challenge of handling long-range dependencies and multi-file contexts~\cite{wang2025position,wang2026fasa,wang2026swepruner,dai2026cedar,hu2025beyondemotion,li2026wrote}. However, most evaluations focus on isolated snippets or functions and do not capture repository-level complexity. Meanwhile, repository-level techniques have advanced rapidly across code generation~\cite{shrivastava2023repofusion,zhang2024codeagent,shi2024code,shi2024between,hu2026line}, translation~\cite{wang2024repotransbench,wang2025evoc2rust}, issue resolution~\cite{jimenez2024swe,chen2025swe}, and optimization~\cite{wang2026effiskill}, with retrieval-augmented~\cite{RepoCoder,ma2024understand,cheng2026resolvingrobustnessprecisiontradeofffinancial,cheng2026enhancingfinancialreportquestionanswering,lai2026transformers,shi2026reasoning} and agent-based~\cite{yang2024swe,li2025swe,zhao2026dllm} approaches showing particular promise, yet standardized evaluation for comprehensive repository understanding remains lacking. To address this gap, we propose a repository-level QA benchmark to evaluate LLMs on realistic, multi-file code understanding tasks.

\section{Conclusion}

In this paper, we present \ourbench, a new benchmark for evaluating LLMs on realistic repository-level code questions. 
\ourbench consists of 720 high-quality question–answer pairs spanning 15 Python repositories from SWE-Bench and SWE-Bench-Live, and is designed to capture practical reasoning challenges such as multi-hop dependencies and cross-file context. 
In future work, we intend to extend \ourbench to additional programming languages and dynamically evolving repositories to enable broader and more robust evaluation of AI-assisted code intelligence tools.

\section*{Limitations}
Despite the strengths demonstrated by \ourbench, several limitations remain.
First, \ourbench focuses on Python repositories, which may limit the generalizability of our findings to other programming languages, despite the language-agnostic design of our methods.
Second, our evaluation relies on a combination of LLM-as-Judge and limited-scale human evaluation; although the latter is used to validate the former, both may introduce biases and may not fully capture all nuances of answer quality.
Third, the benchmark is constructed from static code snapshots and does not reflect the dynamics of evolving repositories.
Finally, while we select 15 diverse repositories, they may not cover the full spectrum of software projects, particularly smaller or highly domain-specific codebases.

\section*{Acknowledgments}
This paper is supported by the National Key Research and Development Program of China (Grant No. 2023YFB4503802) and the Natural Science Foundation of Shanghai (Grant No. 25ZR1401175).

\bibliography{ref}

\newpage
\appendix
\section{Benchmark Construction}
\subsection{Distribution of Collecting Issues}
\label{app:collecting-issue}
\begin{figure}[h]
    \centering
    \includegraphics[width=1\linewidth]{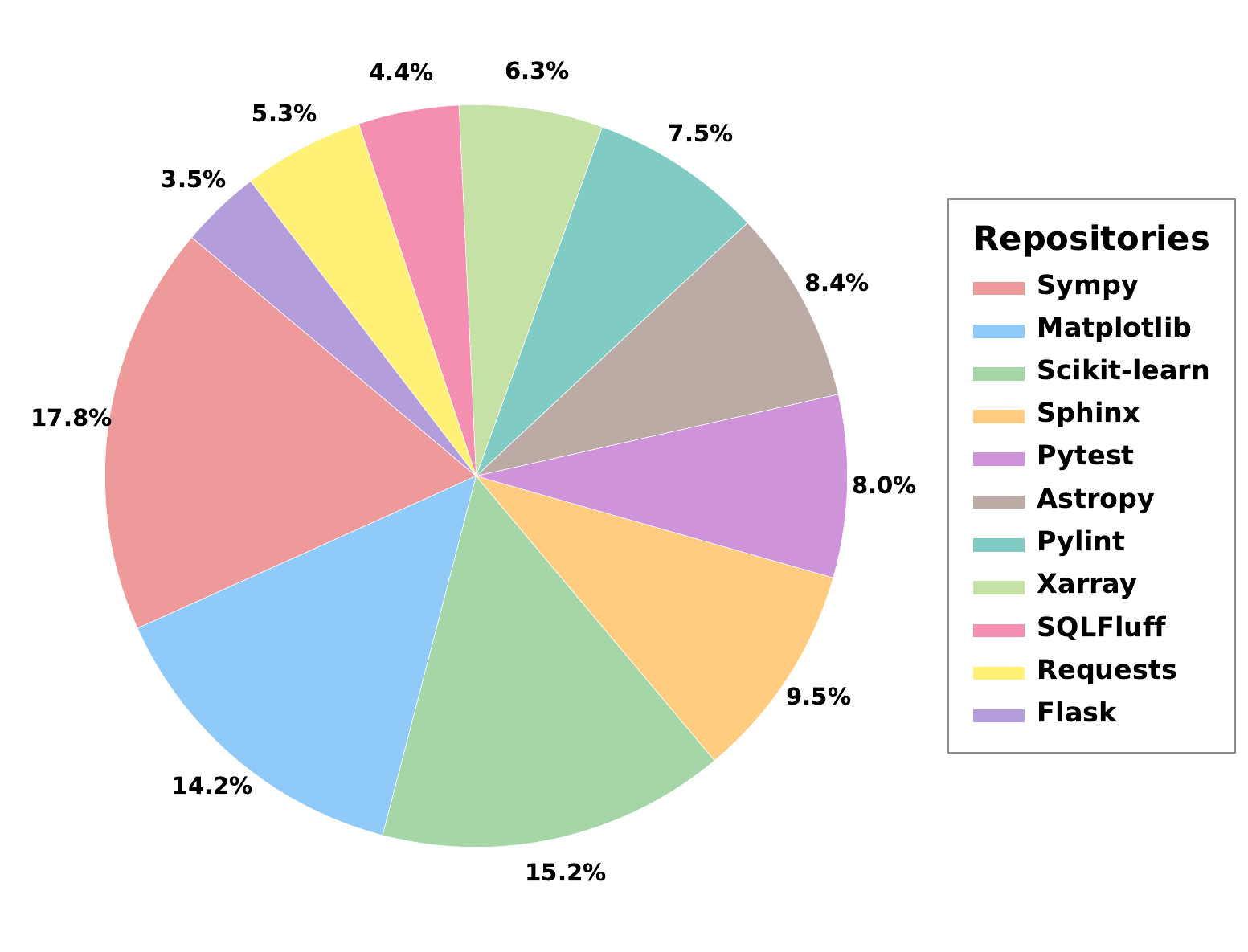}
    \caption{Distribution of collected issues.} 
    \label{fig:issue-distribution}
\end{figure}

The distribution of issues collected across repositories is shown in Figure~\ref{fig:issue-distribution}. Sympy contributed the largest proportion of issues, accounting for 17.8\%, followed by Xarray with 15.2\%. In contrast, Flask and Requests contributed relatively fewer issues, with 3.5\% and 5.3\%, respectively.

\subsection{Code Repository Parsing}
\label{app:repo-structure}
\begin{figure}[ht]
    \centering
    \includegraphics[width=1\linewidth]{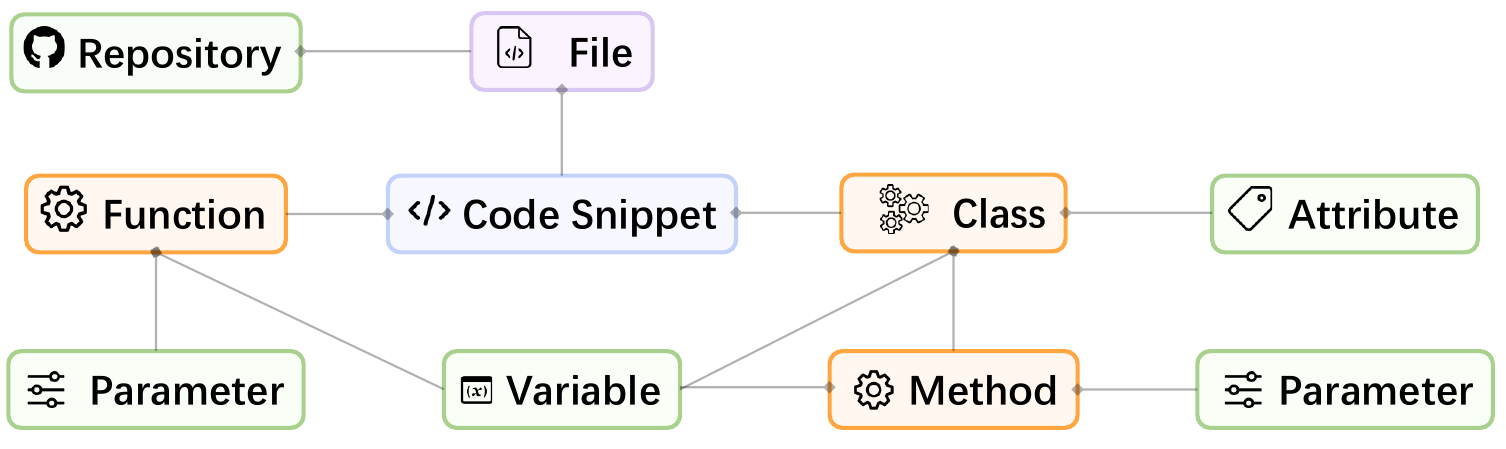}
    \caption{Core elements and their relations extracted from code repositories.}
    \label{fig:parse-overview}
    \vspace{-10pt}
\end{figure}

To extract relevant contextual information, we parse the structure of each repository using \texttt{tree-sitter}\footnote{\url{https://tree-sitter.github.io/tree-sitter/}}, a language-agnostic parsing tool. This process produces a typed graph of core elements and their relationships (Figure~\ref{fig:parse-overview}), where nodes include \textbf{Repository}, \textbf{File}, \textbf{Code Snippet}, \textbf{Class}, \textbf{Method}, \textbf{Attribute}, \textbf{Function}, \textbf{Parameter}, and \textbf{Variable}. Edges represent containment relationships (e.g., Class $\rightarrow$ Method/Attribute, File $\rightarrow$ Code Snippet). Additionally, each function tracks the functions it calls and those that call it, while each file records its imports, revealing inter-file dependencies. Overall, this structure captures both type-level and call-level dependencies, providing the necessary foundation for multi-hop reasoning.

\begin{figure*}[t]
    \centering
    \includegraphics[width=\linewidth]{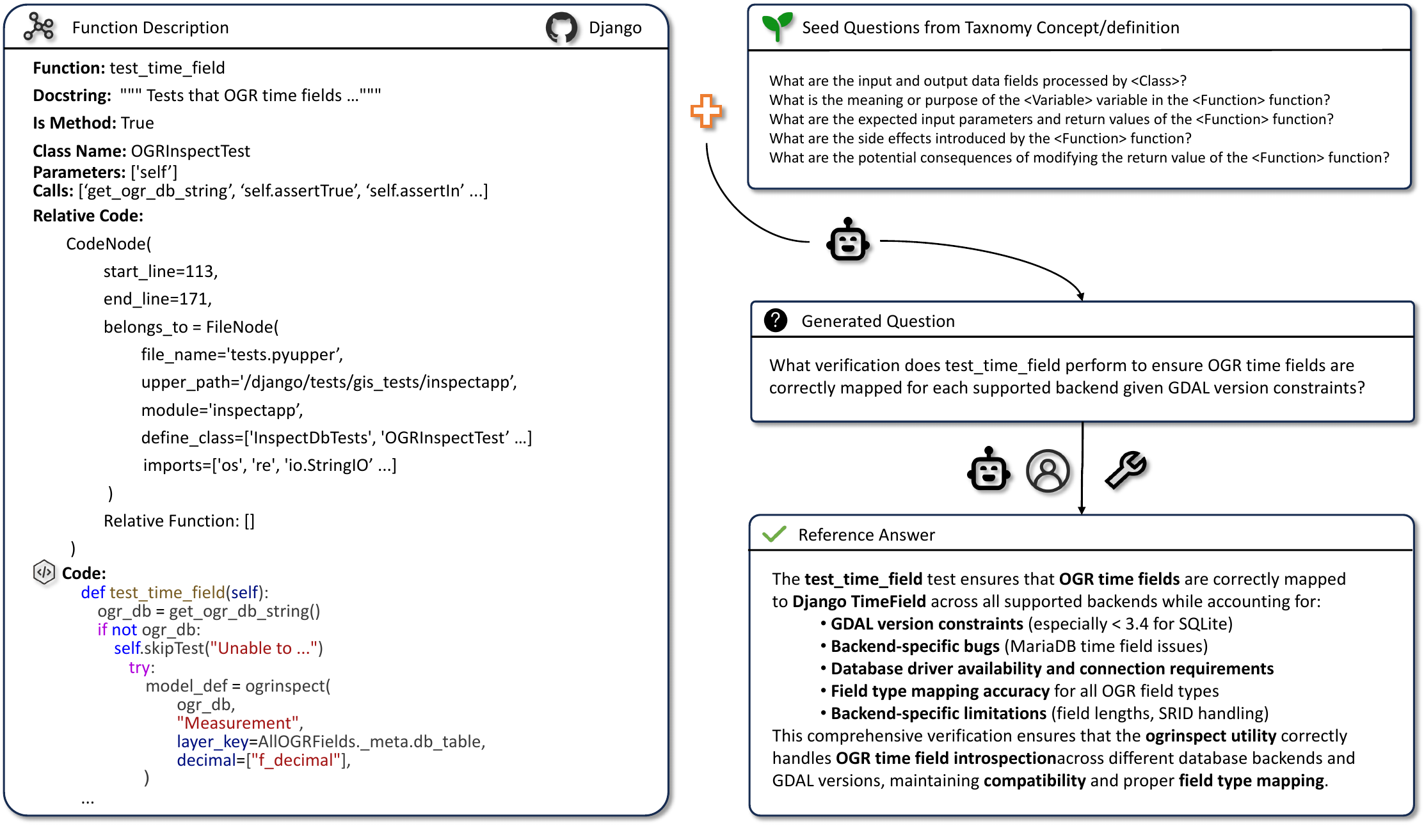}
    \caption{Illustrating question instantiation process.}
    \vspace{-0.5cm}
    \label{fig:instantiation-example}
\end{figure*}

\subsection{Illustrative Example of Question Instantiation}
\label{app:question_example}
Figure~\ref{fig:instantiation-example} illustrates an instance of the question instantiation process. 
The left panel summarizes the focal function and its surrounding context; the top-right panel shows the five seed questions from the selected taxonomy category; the middle-right panel depicts how the LLM synthesizes a single, non-compound question tailored to the target function; and the bottom-right panel presents the curated reference answer with repository-grounded constraints (e.g., version guards and backend differences). 
All structural elements and the complete seed set are provided to the LLM to support candidate question generation.

\begin{table*}[t]
    \centering
    \small
    \resizebox{\linewidth}{!}{
    \begin{tabular}{lccccccc@{}@{}c@{}@{}c@{}c}
    \toprule
    \textbf{Repository} & \textbf{\# Files} & \textbf{\# Classes} & \textbf{\# Functions} & \textbf{\# LOC} & \textbf{\# Questions} & \makecell{\textbf{Avg.}\\\textbf{Words (Q)}} & \makecell{\textbf{Avg.}\\\textbf{Words (A)}} \\
    \midrule
    astropy       & 964   &1,909  & 16,264 & 402,824 & 48 & 32.88  & 175.42 \\
    django        & 2,845 &7,240  & 28,355 & 499,240 & 48 & 30.56  & 362.44 \\
    flask         & 83    &64     & 829    & 18,108  & 47 & 25.66  & 284.62 \\
    matplotlib    & 905   &968    & 9,941  & 266,896 & 48 & 29.85  & 285.46 \\
    pylint        & 2,308 &2,287  & 6,746  & 117,602 & 48 & 26.33  & 268.46 \\
    pytest        & 260   &491    & 5,151  & 100,111 & 48 & 29.10  & 277.19 \\
    requests      & 36    &70     & 598    & 11,248  & 48 & 29.29  & 266.73 \\
    scikit-learn  & 982   &764    & 10,360 & 424,550 & 48 & 29.94  & 283.75 \\
    sphinx        & 743   &1,064  & 6,841  & 142,146 & 48 & 28.48  & 278.02 \\
    sqlfluff      & 392   &2,089  & 1,854  & 145,382 & 48 & 27.52  & 287.48 \\
    sympy         & 1,584 &1,997  & 33,994 & 779,192 & 48 & 26.19  & 268.81 \\
    xarray        & 233   &601    & 7,902  & 186,039 & 48 & 30.88  & 290.96 \\
    conan         & 1,074 &814    & 6,893  & 174,161 & 48 & 28.02  & 507.92 \\
    reflex        & 382   &974    & 3,005  & 98,312  & 48 & 27.52  & 76.46 \\
    streamlink    & 509   &1,190  & 3,671  & 86,593  & 48 & 27.08  & 85.83 \\
    \midrule
    \textbf{Overall} & \textbf{13,300} & \textbf{22,522} & \textbf{142,404} & \textbf{3,452,404} & \textbf{720} & \textbf{28.62} & \textbf{266.64} \\
    \bottomrule
    \end{tabular}
    }
    \vspace{-0.1cm}
    \caption{Statistics of \ourbench. The benchmark comprises 720 questions spanning over 3.4M lines of code across 15 repositories, with answers averaging 266.6 words}
    \label{tab:benchmark}
\end{table*}


\subsection{Detailed Statistics of \ourbench}
\label{app:bench-static}
Table~\ref{tab:benchmark} provides a comprehensive statistical overview of \ourbench across repositories. The 15 repositories span a wide range in size, from 36 to 2,845 files and from 598 to 33,994 functions, offering a thorough representation of repositories of varying scales. Analyzing the benchmark scores in Table~\ref{tab:rq4_main}, we observe that Flask and Requests, each containing fewer than 100 files, achieve the two highest scores, whereas Pylint, with over 2,000 files, attains the lowest score. This indicates a strong correlation between the difficulty of repository comprehension and the inherent complexity of the repository. Additionally, the average question length is 28.62 words, and the average answer length is 266.64 words. For answers, the average edit distance after manual corrections is 42.17 words.

\section{Experimental Details}
\subsection{Method Configurations}
\label{app:experimental-details}

The configurations of all the methods studied are described in the following.

\begin{itemize}[leftmargin=0.5cm]
\setlength{\itemsep}{1pt}
    \item \textbf{Direct Prompting}: The model answers the question without any retrieved context. This baseline evaluates its internal knowledge and serves as a reference for measuring gains from retrieval-augmented generation (RAG) or agent-based methods.
    \item \textbf{Function Chunking RAG}~\cite{wang2024rlcoder}: This approach partitions code based on semantic boundaries, parsing the repository into function-level chunks. The embedding model used is \texttt{voyage-code-3}~\cite{voyageai2024voyagecode3} with 2048 dimensions. A Top‑K setting with K=10 is adopted for the retrieval.
    \item \textbf{Sliding Window RAG}~\cite{RepoCoder}: This method employs a sliding window to extract code snippets by dividing lengthy files into overlapping segments with a chunk size of 500 lines and an overlap of 100 lines. The embedding model used is the same in Function Chunking RAG. A Top‑K setting with K=10 is adopted for the retrieval.   
    \item \textbf{OpenHands}~\cite{wang2024openhands}: This method uses openhands-sdk\footnote{\url{https://github.com/OpenHands/software-agent-sdk}, version 1.1.0}, with the agent set to \texttt{get\_default\_agent} (including tools such as \textit{terminal}, \textit{file\_editor}, \textit{task\_tracker}, \textit{finish}, \textit{think}, etc.).
    \item \textbf{SWE-agent}~\cite{yang2024swe}: This method adapts the original agent, which was designed to handle Issues, to QA tasks, retaining the original agent's action space and memory structure.
    \item \textbf{Commercial Tools}: Tongyi Lingma uses its proprietary model; Cursor runs in its default ``auto'' mode that automatically selects the best model based on the user query with built-in retrieval and orchestration. These commercial tools are evaluated to reflect the performance of current state-of-the-art closed-source solutions. 
\end{itemize}

Unless otherwise specified, OpenHands and SWE-agent is configured with a maximum of \textbf{10} reasoning–action iterations per question. We set all LLM decoding temperatures to 0 to avoid the influence of randomness. All experiments were conducted on a system equipped with an Intel Xeon Gold 6254 CPU and an NVIDIA A100-80G GPU.
The implementations of all methods are publicly available at \url{https://github.com/peng-weihan/SWE-QA-Bench/Scripts/}.

\begin{table*}[t]
    \centering
    \small
    \renewcommand{\arraystretch}{1.05}
    \resizebox{\linewidth}{!}{
\begin{tabular}{l@{\hspace{5mm}}c@{\hspace{3mm}}c@{\hspace{3mm}}c@{\hspace{3mm}}c@{\hspace{3mm}}c@{\hspace{3mm}}c}
\toprule
\multirow{2}[2]{*}{\textbf{Model}} & \multicolumn{5}{c}{\textbf{Evaluation Metrics}} & \multirow{2}[2]{*}{\hspace{2mm} \textbf{Overall}} \\
    \cmidrule{2-6} 
    & \textbf{Correctness} & \textbf{Completeness} & \textbf{Relevance} & \textbf{Clarity} & \textbf{Reasoning} & \\ 
    \midrule
    GPT-5.1                                & 8.75 & 7.52 & 18.02 & 18.12 & 14.02 & 66.43 \\
    \hspace{1em} + Function Chunking RAG   & 10.22 & 10.54 & 18.20 & 18.30 & 16.20 & 73.46 \\
    \hspace{1em} + Sliding Window RAG      & 10.05 & 10.67 & 18.42 & 18.40 & 16.08 & 73.62 \\
    \hspace{1em} + SWE-agent               & 13.54 & 14.07 & 18.66 & 18.50 & 17.02 & 81.79 \\
    \hspace{1em} + OpenHands               & 13.88 & 13.98 & 18.78 & 18.47 & 17.22 & 82.33 \\

    \bottomrule
    \end{tabular}%
    }
    \vspace{-0.1cm}
\caption{Human evaluation results. The result shows high agreement with LLM-as-a-Judge, validating the reliability of automated assessment methods.}
\label{tab:rq2_human}
\end{table*}

\begin{table*}[t]
\centering
\small
\resizebox{1\linewidth}{!}{
\begin{tabular}{l|c c c c c c|c}
\toprule
\textbf{Model} &
\makecell{\textbf{Qwen3-Coder-}\\\textbf{30B-A3B-Instruct}} &
\makecell{\textbf{Qwen3-Coder-}\\\textbf{480B-A35B-Instruct}} &
\makecell{\textbf{Kimi K2}} &
\makecell{\textbf{GLM-4.6}} &
\makecell{\textbf{Gemini 2.5 Pro}} &
\makecell{\textbf{GPT-5.1}} &
\textbf{Average} \\
\midrule
Direct
& 94/63 & 94/65 & 94/34 & 94/76 & 94/51 & 94/87 & 94/63 \\
Function Chunking RAG
& 3,042 / 64 & 3,042 / 106 & 3,042 / 92 & 3,042 / 240 & 3,042 / 82 & 3,042 / 196 & 3,042 / 130 \\
Sliding Window RAG
& 6,302/73 & 6,302/112 & 6,302/110 & 6,302/233 & 6,302/99 & 6,302/201 & 6,302/138 \\
SWE-agent
& 133,293/6,026 & 165,206/4,546 & 154,295/1,398 & 70,870/1,902 & 111,652/6,793 & 122,833/6,100 & 126,026/4,627 \\
OpenHands
& 115,879/2,366 & 108,703/2,128 & 101,594/1,787 & 42,433/2,158 & 74,368/1,116 & 69,192/2,127 & 87,045/1,930 \\

\bottomrule
\end{tabular}}
\vspace{-0.25cm}
\caption{Token Usage per Question (Input/Output). Direct prompting consumes the fewest tokens, while agent-based methods such as SWE-agent and OpenHands require significantly more input and output tokens. This highlights the trade-off between model reasoning complexity and token consumption.}
\vspace{-0.2cm}

\label{tab:token_cost}
\end{table*}

\subsection{LLM-as-Judge Evaluation Protocol}
\label{app:llm-as-judge}

Given a model output and the corresponding gold answer, we use Claude Sonnet 4.5~\cite{claudeSonnet45} as an automatic judge to assess answer quality along five dimensions:
(1) \textit{correctness}, factual accuracy;
(2) \textit{completeness}, coverage of all aspects of the question;
(3) \textit{relevance}, pertinence to the given question;
(4) \textit{clarity}, structural organization and readability; and
(5) \textit{coherence}, logical coherence of the reasoning process.

To improve reliability, each instance is evaluated five times per dimension, with majority voting used to determine dimension-level scores that are then aggregated into a single instance-level score. 
To mitigate bias, candidate systems are anonymized and answer orders are randomly shuffled across trials. 
The judge model is fixed, distinct from all candidate models, and never evaluates its own outputs. 
Overall, this protocol provides a robust assessment of both semantic accuracy and logical coherence. 
The complete LLM-as-Judge prompt is provided in Prompt~\ref{prompt3}.

\section{Additional Experimental Results}
\subsection{Human Evaluation}
\label{sec:human_evaluation}

While the LLM-as-Judge approach offers a scalable evaluation method, it is susceptible to inherent biases. To complement our automated metrics and obtain a more reliable assessment of answer quality, we conduct a human evaluation. Specifically, we recruited three professional software engineers, each with over three years of development experience, who are not co-authors of this paper. For each question, we presented the participants with the reference answer and the answers generated by fives approaches based on the GPT-5.1 model. To ensure fairness, answers were randomized and participants were blind to the generating approach. Each participant was asked to rate the answers on a 20-point scale across the same five dimensions used in our automated evaluation (Section~\ref{sec:dimensions}). This experimental design is consistent with established practices in related research~\cite{geng2024large,liu2023g}.

The results of our human evaluation are presented in Table~\ref{tab:rq2_human}. From the results, although the scores given by human evaluators are generally higher than those from the LLM-as-a-Judge, the overall trends remain highly consistent: both RAG methods show substantial improvement over Direct prompting, and the Agent Framework provides further gains.

\subsection{Cost Analysis} 
In our experiments, we measured the cost of each model across different methods in terms of token usage, as shown in Table~\ref{tab:token_cost}. Compared to direct responses, RAG incurs over ten times more token consumption. Most notably, agent frameworks exhibit an order-of-magnitude increase, consuming approximately 100$\times$ more tokens than direct responses (and 10$\times$ more than RAG).

However, this massive cost discrepancy highlights a critical cost-performance trade-off in repository-level reasoning. Taking GLM-4.6 as an example, employing OpenHands instead of Function Chunking RAG improves Correctness from 9.48 to 11.91 and Completeness from 7.52 to 12.70. This substantial improvement—particularly the nearly 70\% increase in Completeness—indicates that while RAG methods are cost-efficient, they often fail to comprehensively capture all useful context for complex, multi-hop repository queries. Therefore, for practitioners, choosing between RAG and agent frameworks requires carefully balancing budget constraints against the necessity for exhaustive code comprehension. 

Moreover, different LLMs exhibit significant variation in token usage even when using the same agent framework, which is likely related to their intrinsic tool-calling efficiency and planning capabilities.
\section{Case Study}
\label{sec:case_study}

To provide deeper insights into the limitations of current LLMs and reveal common failure patterns in repository-level code comprehension, we conduct a detailed error analysis. Table~\ref{tab:case_study} illustrates a representative case study involving the integration of Sphinx Glossary Term Validation, generated by the best-performing configuration (GPT-5.1 combined with OpenHands).

As shown in the table, the model is capable of identifying high-level system architectures and trigger mechanisms (e.g., \texttt{pytest.mark.sphinx} and the transform pipeline), which earns it respectable scores in Relevance and Clarity. However, it severely lacks analysis granularity. The generated response fails to capture the core components handling the translation (\texttt{i18n}) and completely ignores critical, multi-hop implementation details such as the \texttt{\_NodeUpdater} class and the \texttt{compare\_references()} method. As a result, its Correctness and Completeness scores are extremely low. 

This case study exemplifies a prevalent issue: while modern LLM agents can grasp surface-level code structures, they frequently struggle to navigate and extract precise, deep dependency chains across multiple files. To further support benchmark users and practitioners, we have collected and annotated 100 typical error cases, which are available in our anonymous repository.

\begin{table*}[t]
\centering
\small 
\begin{tabular}{p{0.95\textwidth}} 
\toprule
\textbf{Case Study: Error Analysis of Sphinx Glossary Term Validation} \\
\midrule
\textbf{Question:} \textit{"What is the integration mechanism between the test function that validates glossary term reference consistency and the translation processing transform system?"} \\
\midrule
\textbf{Reference Answer:} \\
The mechanism is connected through the Sphinx docutils transform pipeline. It includes:
\begin{itemize}[leftmargin=*, nosep]
    \item \textbf{Test Setup:} The test is configured using the \texttt{@sphinx\_intl} decorator.
    \item \textbf{Transform Registration:} The \texttt{setup()} in \texttt{sphinx/transforms/i18n.py} adds the \texttt{Locale} transform (priority 20) to the pipeline.
    \item \textbf{Glossary Validation:} The \texttt{Locale} transform's \texttt{\_NodeUpdater.update\_pending\_xrefs()} specifically handles \texttt{refdomain='std'} and \texttt{reftype='term'} cases. It compares the number of glossary term references between the original and translated texts using \texttt{compare\_references()}. If there is a mismatch, a warning of type \texttt{'inconsistent\_references'} is generated.
\end{itemize} \\
\midrule
\textbf{Error Response Analysis} \\
\textbf{1. Insufficient Analysis Granularity}
\begin{itemize}[leftmargin=*, nosep]
    \item \textbf{Issue:} In this specific response, the model stays at a high-level system architecture level (e.g., mentioning \texttt{SphinxTransformer}, \texttt{app.add\_transform()}, and \texttt{SphinxTestApp}), completely ignoring the core components that handle translation (i18n) and consistency validation.
    \item \textbf{Missing Details:} The generated answer lacks descriptions of the \texttt{Locale} transform, \texttt{\_NodeUpdater} class, and key methods like \texttt{update\_pending\_xrefs()} and \texttt{compare\_references()}. It also fails to mention the warning type generated, such as \texttt{'inconsistent\_references'}.
\end{itemize}
\vspace{0.2em}
\textbf{2. Right Direction, Lacking Depth}
\begin{itemize}[leftmargin=*, nosep]
    \item \textbf{Issue:} The model correctly points out that the test is indirectly triggered by \texttt{pytest.mark.sphinx} and \texttt{app.build()} through the transform pipeline (which earns it high clarity and relevance scores). However, at the implementation level, it wrongly attributes the responsibility to a general \texttt{SphinxDomains} instead of the specialized i18n pipeline.
    \item \textbf{Result:} Due to the lack of specific class names and method-level evidence in this response, the correctness score (5/20) and completeness score (4/20) evaluated by the judge are extremely low.
\end{itemize} \\
\bottomrule
\end{tabular}
\caption{A representative case study of an incorrect response generated by GPT-5.1 paired with OpenHands. Although the model successfully identifies high-level interactions, its specific answer completely misses the critical, multi-hop implementation details required for repository-level comprehension.}
\label{tab:case_study}
\end{table*}
\section{Prompts Used in \ourbench}
\label{app:prompt}
\newcounter{prompt}

\refstepcounter{prompt}
\begin{tcolorbox}[
    enhanced,
    colframe=black,
    colback=white,
    coltitle=white,
    colbacktitle=black,
    title=Prompt~\theprompt: Extracting code questions from issues,
    boxrule=0.8pt,
    fonttitle=\mdseries\small,
    fontupper=\ttfamily\scriptsize,
    rounded corners,
    breakable,
    listing only,
    width=\columnwidth,         
    listing options={
        breaklines=true,
        basicstyle=\ttfamily\tiny,
        breakatwhitespace=true,
        columns=flexible
    }
]
\label{prompt1}

You are given a GitHub issue from the \{REPOSITORY\_NAME\} repository. Extract or rewrite it into one or more \\
\textbf{short, clear, concise questions} about understanding the \{REPOSITORY\_NAME\} codebase, APIs, or system design. 

\textbf{Rules:} \\
1. Only include questions answerable by code, logic, or documentation. \\
2. Ignore bug reports, environment issues, or problems that require fixing code. \\
3. Each question should ideally be $\leq 20$ words. 

\textbf{IMPORTANT:} \\
- Be STRICT in quality control: if the issue doesn't contain meaningful questions about code understanding, return an empty questions array. \\
- It's better to return no questions than to generate low-quality or irrelevant questions. \\
- Only extract questions that genuinely help understand the \{REPOSITORY\_NAME\} codebase, APIs, or system design. 

\textbf{Input:}\\
GitHub issue from \{REPOSITORY\_NAME\} repository: \\
Title: \{ISSUE\_TITLE\} \\
Body: \{ISSUE\_BODY\} 

\textbf{Output JSON format:} \\
\texttt{\{ } \\
\texttt{\ \ issue\_number: <number>, } \\
\texttt{\ \ questions: [\{question: <string>\}\ldots ]} \\
\texttt{\} }
\end{tcolorbox}

\refstepcounter{prompt}
\begin{tcolorbox}[
    enhanced,
    colframe=black,
    colback=white,
    coltitle=white,
    colbacktitle=black,
    title=Prompt~\theprompt: Generating repository-level code questions,
    boxrule=0.8pt,
    fonttitle=\mdseries\small,
    fontupper=\ttfamily\scriptsize,
    rounded corners,
    breakable,
    listing only,
    width=\columnwidth,         
    listing options={
        breaklines=true,
        basicstyle=\ttfamily\tiny,
        breakatwhitespace=true,
        columns=flexible
    }
]
\label{prompt2}
You are an expert software research assistant. \\
Given: \\
1. A function/class description extracted from a software repository. \\
2. A list of seed questions as candidates from the \{CATEGORY\} category. \\

Task: \\
1. Based on the seed questions and the function/class description, generate **one single question** that is: \\
- As difficult and complex as possible, \\
- Requires multi-hop reasoning or deep technical understanding, \\
- Not answerable by simple retrieval or direct lookup (i.e., not solvable by basic RAG methods), \\
- Clearly related to the function/class description, \\
- Technically precise and detailed, \\
- Reflects the style and intent of the original seed questions but goes significantly deeper, \\
- **Must not be a compound question** (e.g., no use of !and!, !or!, or comma-based subquestions), \\
- **Must be not too long and syntactically simple**, \\
- **Must be specific to the \{CATEGORY\} category**. \\

2. The question should encourage advanced analysis, integration of multiple concepts, or insight beyond surface-level information. \\
3. Output only the single refined question without additional explanation or commentary. \\

Input: \\
1. Function/Class Description: \\
\{DESCRIPTION\} \\
2. Seed Questions from \{CATEGORY\}: \\
\{SEED\_QUESTIONS\}

\end{tcolorbox}

\refstepcounter{prompt}
\begin{tcolorbox}[
    enhanced,
    colframe=black,
    colback=white,
    coltitle=white,
    colbacktitle=black,
    title=Prompt~\theprompt: LLM-as-a-Judge,
    boxrule=0.8pt,
    fonttitle=\mdseries\small,   
    fontupper=\ttfamily\scriptsize,   
    rounded corners,
    breakable,
    listing only,
    width=\columnwidth,             
    listing options={
        breaklines=true,
        basicstyle=\ttfamily\tiny,
        breakatwhitespace=true,
        columns=flexible
    }
]
\label{prompt3}
You are a STRICT and RIGOROUS evaluator. You must rate the candidate answer STRICTLY against the reference answer. Be CONSERVATIVE with high scores - only award high scores (16-20) when the candidate answer is truly excellent and closely matches the reference answer in quality and content.
\\

CRITICAL EVALUATION PRINCIPLES:\\
1. Compare the candidate answer DIRECTLY with the reference answer point by point\\
2. Any deviation, omission, or inaccuracy should result in score reduction\\
3. High scores (16-20) should be RARE - reserve them only for answers that are nearly perfect\\
4. Be strict about factual accuracy - even minor errors should lower the correctness score\\
5. Missing key points from the reference answer should significantly reduce completeness score\\
6. Vague or imprecise language should lower clarity scores\\
7. When in doubt between two score ranges, choose the LOWER one\\

Evaluation Criteria and Scoring Guidelines (each scored 1 to 20, total score 100):\\
        1. Correctness (STRICT - penalize any inaccuracies):\\
            20 — ONLY if completely correct with ALL core points and details accurate, matching reference answer precisely\\
            16-19 — Mostly correct but must have only TRIVIAL inaccuracies; any noticeable error reduces to 15 or below\\
            12-15 — Partially correct; has some errors or omissions that affect understanding; main points may be accurate but details are wrong\\
            8-11 — Several errors or ambiguities that significantly affect understanding of core information\\
            4-7 — Many errors; misleading or fails to convey key information correctly\\
            1-3 — Serious errors; completely wrong or misleading\\
        2. Completeness (STRICT - penalize missing information):\\
            20 — ONLY if covers ALL key points from reference answer without ANY omission; must match reference in depth\\
            16-19 — Covers most key points but missing some non-trivial information; minor omissions are acceptable\\
            12-15 — Missing several important key points; content is noticeably incomplete compared to reference\\
            8-11 — Important information largely missing; content is one-sided or superficial\\
            4-7 — Covers very little relevant information; seriously incomplete\\
            1-3 — Covers almost no relevant information; completely incomplete\\
        3. Relevance (STRICT - penalize off-topic content):\\
            20 — ONLY if content is fully focused on question topic with NO irrelevant information whatsoever\\
            16-19 — Mostly focused but may have minor peripheral information; any significant off-topic content reduces score\\
            12-15 — Generally on topic but contains some off-topic content that detracts from answer\\
            8-11 — Topic not sufficiently focused; contains considerable off-topic or tangential content\\
            4-7 — Content deviates from topic; includes excessive irrelevant information\\
            1-3 — Majority of content irrelevant to the question\\
        4. Clarity (STRICT - penalize unclear expression):\\
            20 — ONLY if language is exceptionally fluent, clear, and precise; very easy to understand without any ambiguity\\
            16-19 — Mostly fluent and clear but may have minor unclear points; any significant ambiguity reduces score\\
            12-15 — Generally clear but some expressions are unclear or not concise; may require effort to understand\\
            8-11 — Expression somewhat awkward; has ambiguity or lacks fluency that hinders understanding\\
            4-7 — Language obscure; sentences are not smooth; significantly hinders understanding\\
            1-3 — Expression confusing; very difficult to understand\\
        5. Coherence (STRICT - penalize weak logic):\\
            20 — ONLY if the reasoning is exceptionally clear, logical, and well-structured; argumentation is excellent and matches reference quality\\
            16-19 — The reasoning is clear and logical with solid argumentation; minor logical gaps may exist\\
            12-15 — The reasoning is fairly reasonable but has noticeable logical jumps or organization issues\\
            8-11 — The reasoning is average; has logical jumps or organization problems that affect understanding\\
            4-7 — The reasoning is unclear; lacks logical order; difficult to follow\\
            1-3 — No clear reasoning; logic is chaotic\\
\\
INPUT:\\
    Question:\{question\}\\
    Reference Answer:\{reference\}\\
    Candidate Answer:\{candidate\}\\
\\
OUTPUT:\\
    Please output ONLY a JSON object with 5 integer fields in the range [1,20], corresponding\\
    to the evaluation scores:\\
        \{\\
        "correctness": <1-20>,\\
        "completeness": <1-20>,\\
        "relevance": <1-20>,\\
        "clarity": <1-20>,\\
        "cohenrence": <1-20>\\
        \}\\
SCORING INSTRUCTIONS:\\
- Read the reference answer carefully and identify ALL key points, details, and structure\\
- Compare the candidate answer systematically against the reference answer\\
- For each criterion, start with a conservative score and only increase if the candidate truly deserves it\\
- If the candidate answer is significantly shorter, less detailed, or less precise than the reference, reduce scores accordingly\\
- If the candidate answer contains information not in the reference (unless it's clearly relevant and accurate), consider reducing relevance score\\
- When scoring, ask yourself: "Does this candidate answer match the quality and completeness of the reference answer?" If not, reduce scores\\
- Average or mediocre answers should receive scores in the 8-15 range, not higher\\
- Only truly excellent answers that closely match the reference should receive 16-20 scores\\

REQUIREMENT:\\
    No explanation, no extra text, no formatting other than valid JSON. Be strict and conservative with your scores.\\
\end{tcolorbox}

\end{document}